\documentclass[dvipsnames]{article} %
\usepackage{colm2024_conference}

\usepackage{booktabs}
\usepackage{enumitem}
\usepackage{wrapfig}
\usepackage{algorithm}
\usepackage{algpseudocode}
\usepackage{graphicx}
\usepackage{subfigure}
\usepackage[misc]{ifsym}

\usepackage{microtype}
\usepackage{amsmath}
\usepackage{colortbl}
\usepackage[utf8]{inputenc}
\usepackage[T1]{fontenc}
\definecolor{lightgray}{rgb}{0.9,0.9,0.9}
\usepackage{caption}
\usepackage{subcaption}
\usepackage{setspace}
\usepackage{url}
\usepackage{multirow}
\usepackage{tabularx}
\usepackage{blindtext}
\usepackage{pgfplots}
\pgfplotsset{compat=1.18} 
\usepackage{tikz}
\usetikzlibrary{er,positioning,bayesnet}
\usepackage{makecell}
\usepackage{tipa}
\usepackage{siunitx}
\usepackage{nicefrac}
\usepackage{listings}
\usepackage[raster,skins, most]{tcolorbox} %
\usepackage{xltabular}
\usepackage{adjustbox}
\usepackage{xurl}
\usepackage{rotating}
\usepackage[normalem]{ulem}

\usepackage{fontawesome}

\useunder{\uline}{\ul}{}

\usepackage{amsmath,amsfonts,bm}

\def\eqref#1{equation~\ref{#1}}

\def\1{\bm{1}}

\DeclareMathAlphabet{\mathsfit}{\encodingdefault}{\sfdefault}{m}{sl}
\SetMathAlphabet{\mathsfit}{bold}{\encodingdefault}{\sfdefault}{bx}{n}

\renewcommand{\ghlink}{https://github.com/Alibaba-NLP/DeepResearch}

\usepackage{makecell}
\usetikzlibrary{tikzmark}
\makeatletter
\newcommand*\myfontsize{%
  \@setfontsize\myfontsize{7}{8}%
}
\makeatother

\usepackage{fontawesome}   
\newcommand{\fleaf}{\textsuperscript{\fontsize{6pt}{6pt}\selectfont \faLeaf}}

\definecolor{uclablue}{RGB}{159, 195, 224}

\definecolor{uclagold}{RGB}{255, 240, 180}

\definecolor{aliceblue}{RGB}{255, 238, 241}

\definecolor{cadmiumgreen}{rgb}{0.0, 0.42, 0.24}

\definecolor{myred}{rgb}{0.7, 0.3, 0.0}
\definecolor{myblue}{rgb}{0.2, 0.3, 0.6}
\definecolor{babygreen}{rgb}{0.85, 0.97, 0.85}

\definecolor{purple1}{RGB}{126, 107, 196}
\definecolor{purple2}{RGB}{199, 158, 207}
\definecolor{purple3}{RGB}{214, 200, 255}
\definecolor{purple4}{RGB}{254, 240, 255}

\definecolor{deepblue}{RGB}{48, 58, 82}

\newcommand{\symboletongyi}{\raisebox{0pt}{~\includegraphics[scale=0.012]{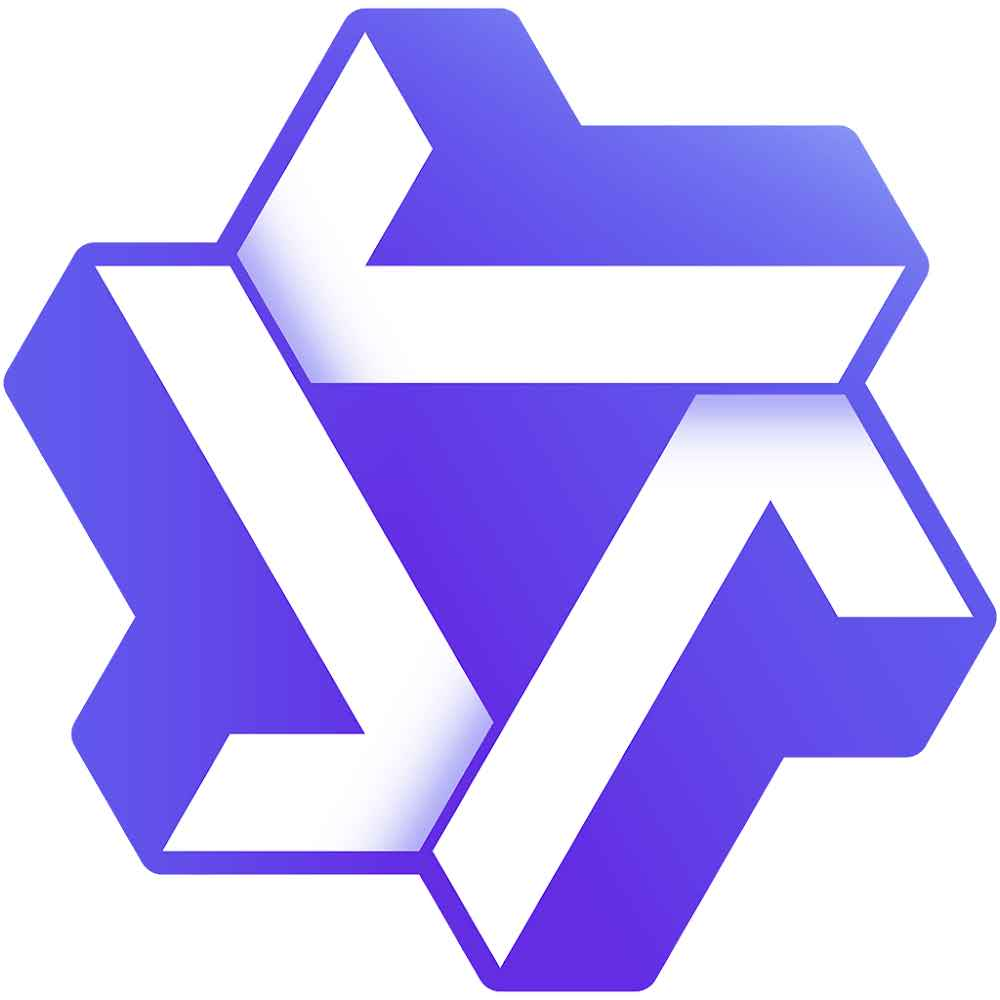}}~}

\definecolor{deepPurple}{HTML}{330066}

\definecolor{uclablue_old}{rgb}{0.15, 0.45, 0.68}
\hypersetup{
    breaklinks,
    citecolor=uclablue_old,
    colorlinks=true,
}

\newtcolorbox{mybox}[2][]
  {colback = black!5!white, colframe = black!75!black, fonttitle = \bfseries,
    colbacktitle = black!100!black, enhanced, before upper={\fontsize{8}{11}\obeyspaces\obeylines\selectfont}, fontupper=\selectfont,
    attach boxed title to top left={yshift=-2.2mm,xshift=4mm},
    title=#2,#1}

\author{%
\small{Runnan Fang$^{*}$, Shihao Cai$^{*}$, Baixuan Li$^{*}$, Jialong Wu$^{*}$$^\fleaf$$^{(\textrm{\Letter})}$, Guangyu Li\thanks{Equal contributors. $^\fleaf$Jialong Wu (wujialongml@gmail.com) is project leader. $^{\textrm{\Letter}}$ yongjiang.yj@alibaba-inc.com}, Wenbiao Yin, \\ 
Xinyu Wang,  Xiaobin Wang, Liangcai Su, Zhen Zhang, Shibin Wu, Zhengwei Tao, Yong Jiang$^{(\textrm{\Letter})}$, Pengjun Xie, Fei Huang, Jingren Zhou}%
  \\[1em]               
  {\fontsize{10pt}{11pt}\selectfont          
Tongyi Lab\symboletongyi, Alibaba Group}\\
}

\usepackage{caption}
\usepackage{subcaption}

\begin{document}

\title{AgentScaler: Towards General Agentic Intelligence \\via Environment Scaling}

\title{%
\raisebox{-2.0em}{
  \parbox[t]{0.35in}{\includegraphics[width=0.6in]{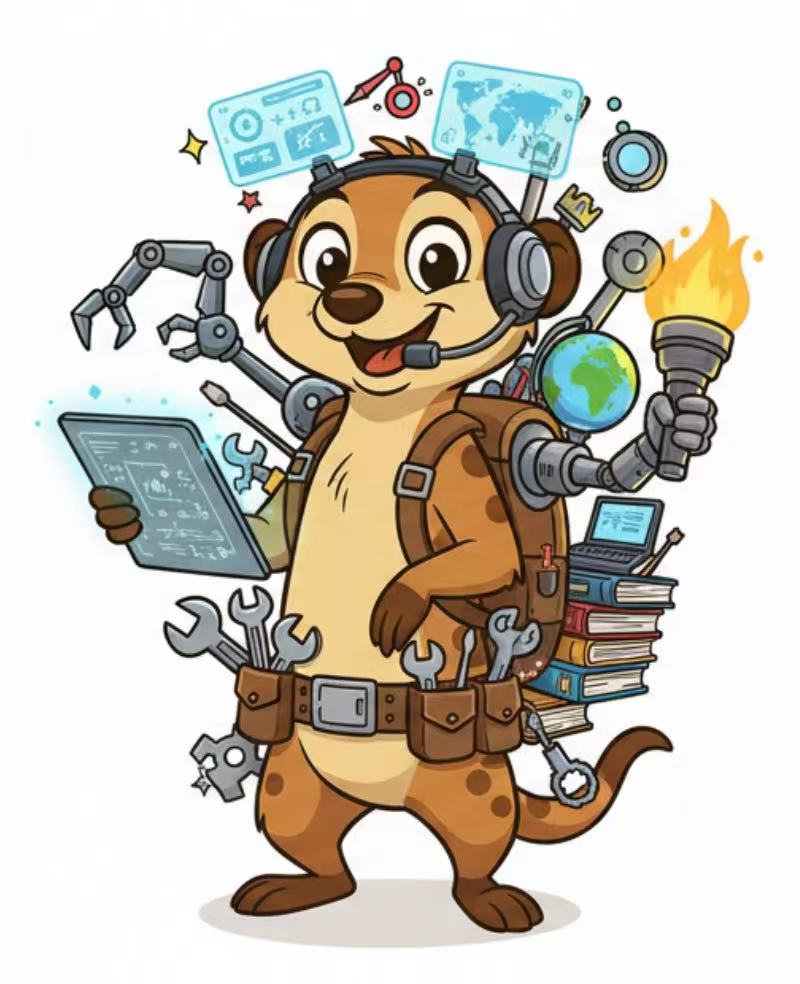}} 
  }
\begin{tabular}[t]{l} 
  \parbox[t]{0.8\textwidth}{\centering 
    Towards General Agentic Intelligence \\
    via Environment Scaling
  }
\end{tabular}
}

\maketitle


\maketitle

\begin{abstract}
  Advanced agentic intelligence is a prerequisite for deploying Large Language Models in practical, real-world applications.
Diverse real-world APIs demand precise, robust function-calling intelligence, which needs agents to develop these capabilities through interaction in varied \textit{environments}.
The breadth of function-calling competence is closely tied to the diversity of environments in which agents are trained.
In this work, we scale up environments as a step towards advancing general agentic intelligence. 
This gives rise to two central challenges: \textit{(i)} how to scale environments in a principled manner, and \textit{(ii)} how to effectively train agentic capabilities from experiences derived through interactions with these environments.
To address these, we design a scalable framework that automatically constructs heterogeneous environments that are fully simulated, systematically broadening the space of function-calling scenarios.
We further adapt a two-phase agent fine-tuning strategy: first endowing agents with fundamental agentic capabilities, then specializing them for domain-specific contexts.
Extensive experiments on agentic benchmarks, $\tau$-bench, $\tau^2$-Bench, and ACEBench, demonstrate that our trained model, \textbf{AgentScaler}, significantly enhances the models’ function-calling capability.

\end{abstract}

\section{Introduction}
Function calling empowers language agents to interface with the real world~\citep{qin2023toolllm,chen2024agent,qin2024tool,schick2023toolformer}.
Yet, their progress is fundamentally constrained by the scarcity of agentic data\footnote{In this paper, the terms ``function-calling'', ``tool'', ``API'', ``MCP'' are used interchangeably; ``agentic data'' refers to trajectories involving such interactions.}, \textit{i.e.}, trajectories generated by autonomous agents interacting with environments via explicit action executions, namely, tool calls ~\citep{zhou2023agents,liu2024toolace}.
The community has gradually transitioned from the era of raw corpora and human-curated data to the emerging \textbf{era of experience}~\citep{silver2025welcome,wu2025webdancer,li2025websailor,tao2025webshaper,geng2025webwatcher}.
Crucially, language agents must experience these interactions themselves in a predefined environment, which makes both data collection and reliable supervision highly challenging.

Several approaches have been attempted to generate synthetic agentic data.
Broadly, previous methods fall into two categories.
The first category follows a reverse paradigm, in which user queries are generated to match each assistant function call observed at every interaction turn~\citep{magnet}, though the resulting trajectories may exhibit limited realism.
The second category follows a forward paradigm, which we refer to as simulated agent–human interplay~\citep{button,apigen,apigenmt,tau2,zeng2025boosting}. 
Such generated trajectories, however, may lack naturalness. In this category, a high-level user intent is first formulated to necessitate agent interaction. Agentic data is then constructed in a top-down manner based on this intent through human–agent interplay. Yet, the environment is not scalable: the absence of automated environment construction hinders large-scale deployment and inevitably entails some degree of manual intervention.

To address these challenges, we pursue the advancement of general agentic intelligence via systematic environment scaling.
Our approach follows a principled two-stage pipeline: \textit{(i)} \textbf{\textit{fully simulated environment construction and scaling}}, responsible for establishing and expanding diverse agentic scenarios, and \textit{(ii)} \textbf{\textit{agent experience learning}}, which exploits these environments to foster generalizable intelligence.

In designing environment construction and scaling, we follow the principle that the core of an agent lies in its capacity for environment interaction, with each environment instantiated as a \texttt{read}–\texttt{write} database~\citep{tau2,zeng2025boosting}.
Specifically, we collect a broad spectrum of APIs and organize them into domains using community
detection, where each domain represents an environment aligned with a specific database structure.
Then, we instantiate tools as executable code, thereby achieving programmatic materialization that enables direct operations on the underlying database structures.
Finally, we sample from the domain‑specific tool graph to generate parameters for the tool sequences and initialize the corresponding database state. We then integrate these components into an overall user intent, grounding tool executions directly on the database. This design enables verifiability at both the environment level and the tool‑argument response level.

For learning from agent experience, our focus is on training the agent’s ability to perform tool calls and to respond effectively to users~\citep{ye2025feedback,su2025learn}.
We begin by performing simulated human–agent interactions on the constructed agentic tasks~\citep{apigenmt}, thereby collecting trajectories that serve as the agent's experience and perform strict filtering.
To facilitate the acquisition of this capability, we adopt a two‑stage agent experience learning framework: in stage 1, the agent acquires fundamental tool‑calling skills across general domains; in stage 2, it is further trained within target vertical domains using domain‑specific scenarios, enabling smoother and more context‑aligned development of agentic capabilities.

Extensive experiments on agentic benchmarks, $\tau$-bench~\citep{taubench}, $\tau^2$-Bench~\citep{tau2}, and ACEBench~\citep{chen2025acebench} show the effectiveness of our pipeline and trained models.
Based on the above pipeline, we train our family of \textbf{AgentScaler} models (4B, 8B, 30B-A3B), built upon the Qwen-3~\citep{qwen3} series. 
At each comparable scale (4B, 8B), our models achieve state-of-the-art performance.
Notably, AgentScaler-30-A3B sets a new state-of-the-art with significantly fewer parameters, delivering results on par with existing 1T-parameter models and leading closed-source systems.
We also provide a systematic analysis covering model generalization, stability, and the long-horizon tool-calling challenge, offering key insights into the development of general agentic intelligence.

\section{Environment Build and Scaling}
\noindent \textcolor{mypurple}{\textbf{Design Principal}}
In essence, any function call can be interpreted as a \texttt{read}–\texttt{write} operation over an underlying environmental database $\mathcal{D}$~\citep{guo2025stabletoolbench}.
Specifically, each function $func$ can be assigned an operator type, $\mathrm{op}(func) \in \{\texttt{read}, \texttt{write}\}$, where \texttt{read}-type function perform queries over $\mathcal{D}$ (\textit{e.g}., retrieval, inspection, monitoring), while \texttt{write}-type tools induce state transitions in $\mathcal{D}$ (\textit{e.g.}, modification, generation, actuation). 
Under this abstraction, a tool response is equivalent to evaluating the induced operator on $\mathcal{D}$, \textit{i.e.}, $\text{API}(func, \alpha) \equiv \mathrm{op}(func)(\alpha; \mathcal{D})$, where the symbol $\alpha$ denotes the input arguments provided to a function call.
Furthermore, let $\mathcal{T}_d$ denote the set of tools within domain $d$. 
Tools in the same domain typically exhibit structurally similar read–write patterns, which can be captured by a common database schema $\mathcal{S}_k$. 
Consequently, the design problem reduces to defining a partition of the tool space into domains $\{\mathcal{T}_1\, \ldots,\mathcal{T}_M\}$, and assigning to each domain a database schema $\mathcal{S}_k$, where $\mathcal{S}_k$ specifies the environment for that domain.

\begin{figure}[htbp]
\centering
\includegraphics[width=1\columnwidth]{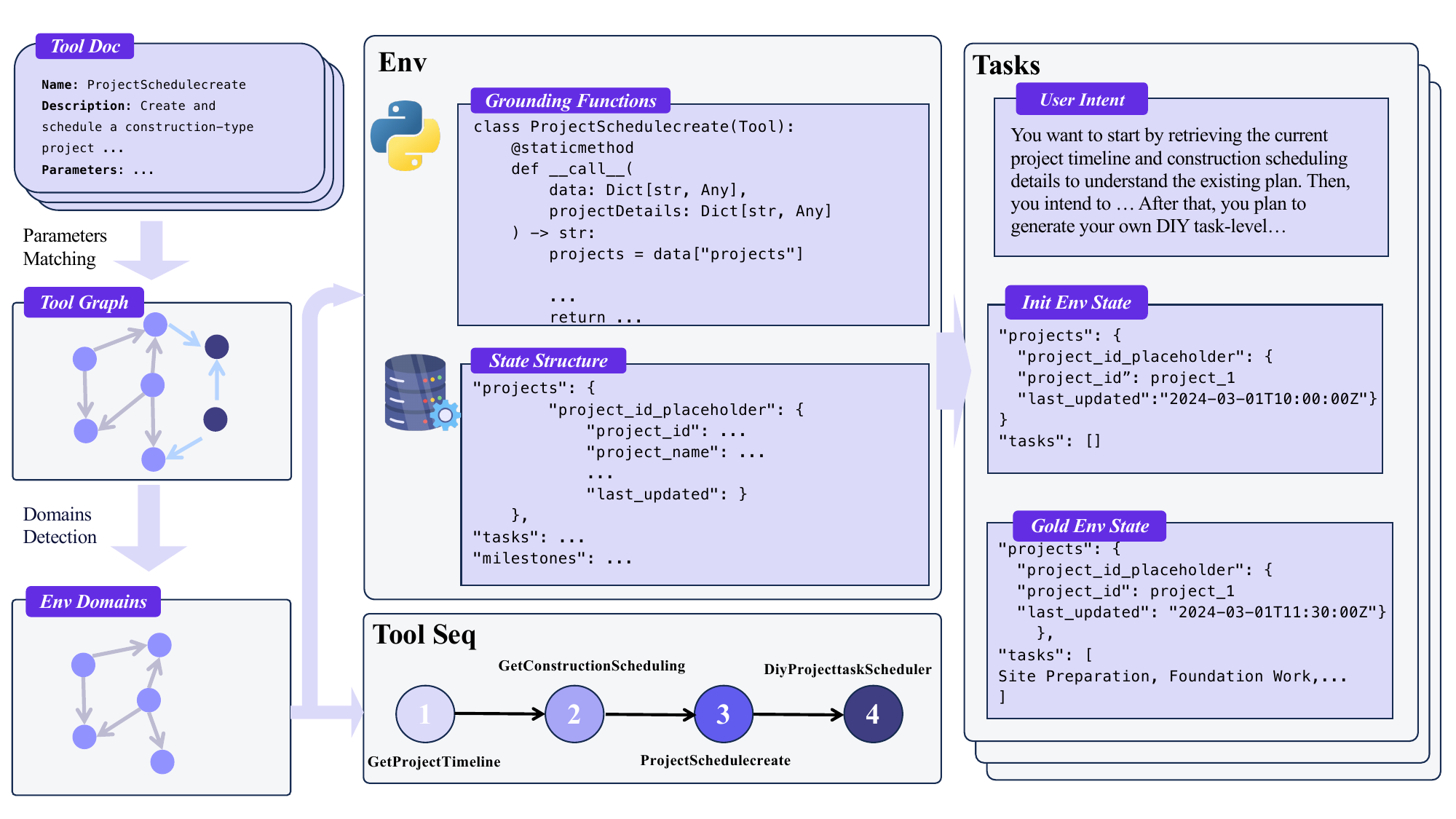}
\caption{The overview of the environment automatic build, and agentic task construction.
}
\label{fig:env_build}
\end{figure}

\subsection{Environment Automatic Build}

Building upon this design principle, we propose a systematic pipeline for leveraging a diverse set of tools as shown in Figure~\ref{fig:env_build}. 
We begin with \textbf{scenario collection}, which gathers a large corpus of real-world tools; proceed to \textbf{tool dependency graph modeling}, which induces well-structured domain partitions and distributions; and finally employ \textbf{function schema programmatic materialization}, which maps tool operations onto database interactions, thereby enabling the construction of the overall environment.

\noindent \textbf{Scenario Collection} We collected more than 30,000 APIs from ToolBench~\citep{qin2023toolllm,guo2024stabletoolbench}, API-Gen~\citep{prabhakar2025apigen} and our internal tool repository.
After applying rigorous filtering, including the removal of low-quality APIs and subsequent refinement, we rewrite some API descriptions to incorporate explicit input–output specifications~\citep{fang2025synworld}.
Building on this, we further constructed tool compositions by systematically exploiting the input–output relationships among APIs.
This process ultimately resulted in API pools $\Theta_{F}$ whose size = $N$ (over 30,000), providing a reliable foundation for subsequent experiments and analysis.

\noindent \textbf{Tool Dependency Graph Modeling} We construct a tool graph in which nodes are tools and edges encode compositional compatibility induced by function parameters.
A tool $func$ consists of a description $P_{func}$ and a list of parameters $P_{func}$.
For a pair of tools, we can extract their respective parameter lists and convert them into vector representations $\phi$ to compute their cos-similarity. If the similarity exceeds a predefined threshold $\tau$, we consider there to be a dependency relationship between the two tools. Accordingly, we insert an edge $E$ between them in our graph.
\begin{equation}
E = \left\{(i, j) \mid \text{sim}(\phi(P_{func_{i}}), \phi(P_{func_{j}})) > \tau,\ i \neq j \right\}
\end{equation}
Domain partitioning then reduces to a graph clustering problem.
We employ Louvain community detection~\citep{blondel2008fast} to identify coherent tool communities that serve as domains. 
For a segmented tool set, since parameter matching relies solely on vectorization and considers only individual parameter information, the overall inter-tool dependencies may be difficult to capture. 
Therefore, for tools within a given domain, we further employ an LLM to systematically examine the dependencies between each pair of tools, thereby further improving the accuracy of edges in the tool graph.
In total, we obtained $M$ domains (exceeding 1,000).

\noindent \textbf{Function Schema Programmatic Materialization}
We first leverage the parameters of all tools within a domain to generate a domain-specific database structure, which serves as the underlying state for subsequent tool operations.
After obtaining the domain-specific tool set and the corresponding database schema in the previous stage, we can formalize each tool in python code, enabling it to perform \texttt{read}–\texttt{write} operations over the database schema. 
Interestingly, when generating database structures and formalizing code within specific domains of $\tau$-bench, we observe through manual inspection that our outputs exhibit a high degree of consistency with the official implementations provided by $\tau$-bench~\citep{taubench}.

\subsection{Agentic Task Construction}
We construct trajectories via forward simulated agent–human interplay, which allows us to fully simulate the environment, the user, and the agent.
The critical step is to synthesize agentic tasks that elicit human tool usage while ensuring that the resulting trajectories remain \textbf{verifiable}.
Concretely, we first initialize an environment state based on the domain-specific database schema, while encouraging as much diversity as possible in the initial state.
Next, we sample logically coherent tool sequences from the domain’s tool graph, specifically by constructing a directed dependency graph over APIs and traversing it to obtain valid sequences. 
Starting from a randomly selected initial node, we conduct a directed walk until either the maximum execution steps are reached or a node with no outgoing edges is encountered. 
This process yields a logically coherent tool sequence.
For each step, we generate the corresponding arguments and perform the actual tool call, grounding the operations directly on the database and continuously tracking the evolving database state.
This procedure enables verifiability at two complementary granularities: \textit{(i)} database-level state consistency and \textit{(ii)} exact matching of tool sequences.

\section{Agent Experience Learning}
We leverage user intent to drive interactions that yield agent experiences, and train the model through a two-phase process.
\begin{figure}[htbp]
\centering
\includegraphics[width=0.9\columnwidth]{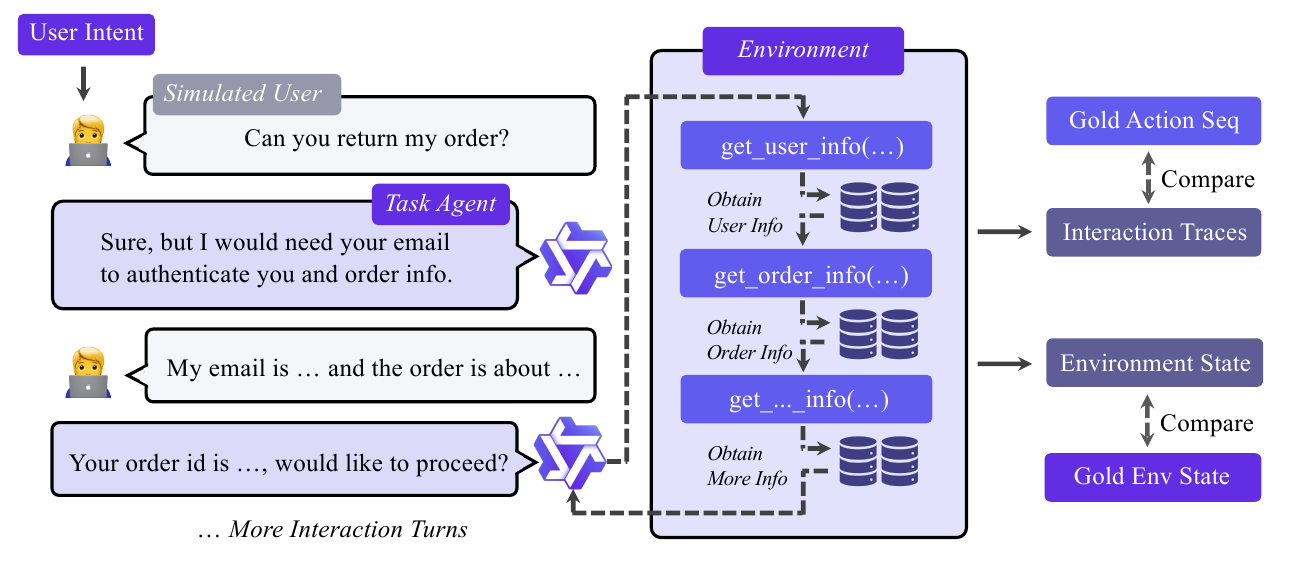}
\caption{The agent interacts with the simulated user and changes the environment state through the generated functions.
\label{fig:infer}
}

\end{figure}
\subsection{Human–Agent Interplay for Experience Collection}

\paragraph{Interplay} Motivated by \cite{taubench}, once we have constructed an agentic task, we proceed to perform human-agent interplay in the environment. 
Specifically, we instantiate a simulated user tasked with fulfilling a given overall intent.
The agent then leverages domain-specific tools to address the user’s needs, continuing the interaction until the simulated user deems the task complete.
This setup enables us to conduct \textbf{end-to-end simulation}, encompassing user simulation, agent, and environment, yielding a highly scalable framework.
Each completed interaction trace constitutes an agent experience, which can subsequently be used for training. 
Importantly, since we possess both the gold tool sequences and arguments for the overall intent and the final environment state, we can apply these as supervision signals for experience filtering.

\paragraph{Filtering} We adopt a three-stage funnel-based trajectory filtering framework consisting of \textit{validity control}, \textit{environment state alignment}, and \textit{function calling exact match}.
\begin{itemize}[itemsep=0.0pt,topsep=0pt]
    \item \textit{\underline{Validity control}}, removes invalid interaction trajectories to ensure well-formed alternating user assistant exchanges. 
    Additionally, we apply an $n$-gram-based filtering procedure to eliminate severely repetitive reasoning segments. In such cases, we discard these data points. 
    \item \textit{\underline{Environment state alignment}} retains only those trajectories whose final database state matches the golden state after the interplay, thereby validating the effectiveness of write operations. The filtering granularity at this stage is the \textbf{database/environment level}.
    \item \textit{\underline{Function calling exact match}} serves as the most stringent filtering stage, where the granularity is the \textbf{tool sequence}.
    Since a tool sequence consisting entirely of read operations without any write operations would cause state-based filtering to fail, we adopt a stricter exact match approach for filtering in such cases.
    A trajectory is preserved only if the sequence of invoked tools and arguments exactly matches the overall intent, ensuring high-fidelity supervision.
\end{itemize}
It is worth noting that we do not filter out trajectories in which tool calls return errors.
Thanks to the aforementioned filtering framework, such trajectories may still accomplish the intended goal despite intermediate failures.
Retaining them in the training data helps improve the robustness of the model.
\subsection{Agentic Experience Learning}
\noindent \textbf{Agentic Fine-tuning}
Given agent-human interplay experience trajectory $
\mathcal{H} = (h_0, a_1, \ldots, a_{n-1}, h_n, a_0)
$, where each human instruction is denoted by $h_t$ at $t$-round interaction, and each assistant turn $a_t$ is decomposed as
$
a_t = (\tau_t, \rho_t, y_t).
$
Here, $\tau_t$ represents the function call tokens, $\rho_t$ the tool response tokens, and $y_t$ the assistant response tokens.
Our training objective is to optimize only the tool calls and assistant responses, while human instructions $h_i$ and tool responses $\rho_t$ are excluded from the loss.
Formally, given an autoregressive model $p_\theta(x_k \mid x_{<k})$, we define the loss as
\begin{equation}
\mathcal{L}(\theta)
= - \frac{1}{\sum_{k=1}^{|\mathcal{H}|} \mathbb{I}[k_i \in \mathcal{T}] }
\sum_{k=1}^{|\mathcal{H}|} \mathbb{I}[x_k \in \mathcal{T}] \cdot
\log \pi_{\theta} \left(x_i \mid x_{<k}\right),
\end{equation}

where $x_k$ denotes the $k$-th token in the trajectory, $\pi_\theta$ is the model distribution, $\mathbb{I}[\cdot]$ is the indicator function, $\mathcal{T}$ is the set of tokens belonging to tool calls $\tau$ or assistant responses $y$.
In practice, all tokens in $\rho_i$ and $h_i$ are masked out from supervision but remain visible in the context $x_{<k}$. 
This ensures that the model conditions on tool responses and human instructions, while gradients are only propagated through assistant-generated tool calls and natural-language responses.
\paragraph{Two-stage Experience Learning} 
In the first phase, the agent is trained to acquire fundamental skills for tool usage and user interaction. 
We focus on general domains where a broad set of tools and tasks are available, allowing the agent to develop a robust understanding of when and how to invoke function calls, as well as how to integrate tool outputs into coherent user-facing responses.
This stage emphasizes breadth and generality, ensuring that the agent builds a versatile foundation of agentic behaviors before domain-specific specialization.
In the second phase, the agent undergoes fine-grained training in vertical domains, where tasks, tools, and user intents exhibit domain-specific characteristics.
By grounding the learning process in realistic scenarios within a target domain, the agent refines its ability to select tools, parameterize calls, and produce responses that are accurate, contextually appropriate, and aligned with domain-specific goals.
This specialization ensures a smoother adaptation of agentic capabilities, enabling the agent to operate effectively in real-world, task-oriented environments.
\section{Experiments}
\subsection{Setup}
\noindent \textbf{Benchmarks}
We evaluate our methods on three established agentic benchmarks: $\tau$-bench, $\tau^2$-Bench, and ACEBench-en. 
For $\tau$-Bench (covering the \texttt{retail} and \texttt{airline} domains) and $\tau^2$-Bench (spanning the \texttt{retail}, \texttt{airline}, and \texttt{telecom} domains), we adopt the pass$\textasciicircum$ 1 metric for evaluation and additionally analyze the trend of pass$\textasciicircum$ k, following the protocols in ~\cite{taubench, tau2}.
For ACEBench-en, we report results across the \texttt{Normal}, \texttt{Special}, and \texttt{Agent} categories, as well as the \texttt{Overall} performance, using the accuracy metric.

\noindent \textbf{Baselines} We compare our trained series models against the following types: \textit{closed-sourced large language model}, including Gemini-2.5-pro~\citep{comanici2025gemini}, Claude-Sonnet-4~\citep{claude}, GPT-o3, GPT-o4-mini~\citep{o3}, and GPT-5 (with thinking)~\citep{gpt5};
\textit{open-sourced large language model}: GPT-OSS-120B-A5B~\citep{gptoss}, Deepseek-V3.1-671B-A37B~\citep{dpskv3},  Kimi-K2-1T-A32B~\citep{team2025kimi}, 
Qwen3-Thinking-235B-A22B~\citep{qwen3}, Seed-OSS-36B~\citep{seedoss}, Qwen-Coder-30B-A3B~\citep{qwencoder}, and xLAM-2 model series~\citep{apigenmt}.

\noindent \textbf{Backbones}
We train the AgentScaler model series by training on Qwen3 models~\citep{qwen3} of varying scales.
Specifically, AgentScaler-4B and AgentScaler-30B-A3B are trained on Qwen3-Thinking-4B-2507 and Qwen3-Thinking-30B-A3B-2507, respectively, while AgentScaler-8B is trained on Qwen3-8B.
\subsection{Experimental Results}
\begin{table}[h]
\small
\centering
\caption{\textbf{Main results} on $\tau$-bench, $\tau^2$-Bench, and ACEBench-en.
}
\label{tab:main_result}
\resizebox{\columnwidth}{!}{%
\begin{tabular}{@{}l|cc|ccc|cccc@{}}
\toprule
&  \multicolumn{2}{c|}{\textbf{ $\tau$-bench}} & \multicolumn{3}{c|}{\textbf{$\tau^2$-Bench}} &\multicolumn{4}{c}{\textbf{ACEBench-en}}  \\ \midrule
\textbf{Model} &\texttt{Retail} & \texttt{Airline} &\texttt{Retail} & \texttt{Airline} & \texttt{Telecom} & \texttt{Normal}& \texttt{Special} & \texttt{Agent} & \texttt{Overall} \\ 
\midrule
\multicolumn{10}{c}{\cellcolor{blue!20}\textbf{\textit{Closed-Source Large Language Models}}}\\
\midrule
Gemini-2.5-pro & 68.7& 44.0& 67.5& 56.0& 27.2& 76.7& 90.0& 63.4& 78.2\\
Claude-Sonnet-4 & 73.9& 40.0& 67.5& 54.0& 47.4& 79.9& 87.3& 42.5& 76.1\\
GPT-o3 & 70.4& 52.0& 80.2& 64.8& 58.2 & 78.3& 86.7& 63.3& 78.2\\
GPT-o4-mini & 70.4& 46.0& 70.2& 56.0& 46.5& 79.9& 84.0& 60.0& 77.9\\
GPT-5-think & 78.3& 44.0& 81.1& 62.6& 96.7& 76.7& 85.3& 32.5& 72.2\\
\midrule
\multicolumn{10}{c}{\cellcolor{blue!30} \textbf{\textit{Open-Source Large Language Models}}} \\
\midrule
GPT-OSS-120B-A5B & 67.8& 49.2& 57.0 & 38.0& 45.6& 79.1& 84.0& 50.8& 76.0\\
Deepseek-V3.1-671B-A37B & 66.1& 40.0& 64.9& 46.0& 38.5& 80.3& 62.0& 40.8& 69.3\\
Kimi-K2-1T-A32B & 73.9& 51.2& 70.6& 56.5& 65.8& 78.9& 81.3& 65.0& 77.4\\
Qwen3-Thinking-235B-A22B & 67.8& 46.0& 71.9& 58.0& 45.6& 72.1& 84.0& 39.1& 70.2\\
\arrayrulecolor{black!20}\midrule
Seed-OSS-36B & 70.4& 46.0& 68.4& 52.0& 41.2& 79.1 & 82.0 & 58.4 & 76.7\\
Qwen-Coder-30B-A3B & 68.7 & 48.0 & 60.5& 42.0& 30.7& 74.0& 41.3& 24.1& 57.5\\
\arrayrulecolor{black!20}\midrule
xLAM-2-8B-fc-r & 58.2 &	35.2 & 55.3 & 48.0 & 11.4 & 58.8 & 0.0 & 5.0 & 34.8 \\
xLAM-2-32B-fc-r & 64.3 & 45.0 & 55.3 & 52.0 & 16.7 & 69.2 & 24.7 & 13.4 & 52.5 \\
xLAM-2-70B-fc-r & 67.1 & 45.2 & 61.4 & 56.0 & 14.0 & 57.1 & 5.3 & 38.4 & 36.5 \\
\arrayrulecolor{black!20}\midrule
Qwen3-Thinking-4B & 59.1& 52.5 & 56.1 & 52.0 & 28.7& 43.3& 84.7& 11.7& 49.5\\
Qwen3-8B  & 45.2& 25.0& 41.2&30.5 & 23.5& 71.4& 75.3& 29.1& 65.9\\
Qwen3-14B  & 45.7& 31.0& 48.0&30.0 &26.9& 66.9& 84.0& 44.2& 68.0\\
Qwen3-Thinking-30B-A3B & 67.8& 48.0& 58.8& 58.0& 26.3& 64.7& 86.7& 42.8& 67.2\\
\midrule
\textbf{AgentScaler-4B} & 64.3& 54.0& 62.3& 56.0& 48.2& 70.3& 76.7& 30.8& 65.9\\
\textbf{AgentScaler-8B}  & 50.4& 42.0& 58.8& 44.0& 45.4& 69.2& 76.7& 44.2& 67.4\\
\textbf{AgentScaler-30B-A3B} & \textcolor{mypurple}{\textbf{70.4}} & \textcolor{mypurple}{\textbf{54.0}} & \textcolor{mypurple}{\textbf{70.2}} & \textcolor{mypurple}{\textbf{60.0}} & \textcolor{mypurple}{\textbf{55.3}} & \textcolor{mypurple}{\textbf{76.7}} & \textcolor{mypurple}{\textbf{82.7}} & \textcolor{mypurple}{\textbf{60.0}} &
\textcolor{mypurple}{\textbf{75.7}} \\
\arrayrulecolor{black}\bottomrule
\end{tabular}
}
\end{table}

\paragraph{Main Results} 
From Table~\ref{tab:main_result}, we observe that closed-source large language models (LLMs) still maintain a clear performance advantage, consistently achieving the highest scores across most domains and benchmarks. This demonstrates the strength of industrial-scale training pipelines and proprietary optimization strategies.
Nevertheless, our proposed AgentScaler achieves a remarkable level of performance given its lightweight parameter scale. Specifically, it surpasses most open-source baselines with fewer than 1T parameters, establishing a new state-of-the-art across $\tau$-bench, $\tau^2$-Bench, and ACEBench-en. 
\begin{wrapfigure}{r}{8cm}
\vspace{-5mm}
\centering
\includegraphics[width=0.35\columnwidth]{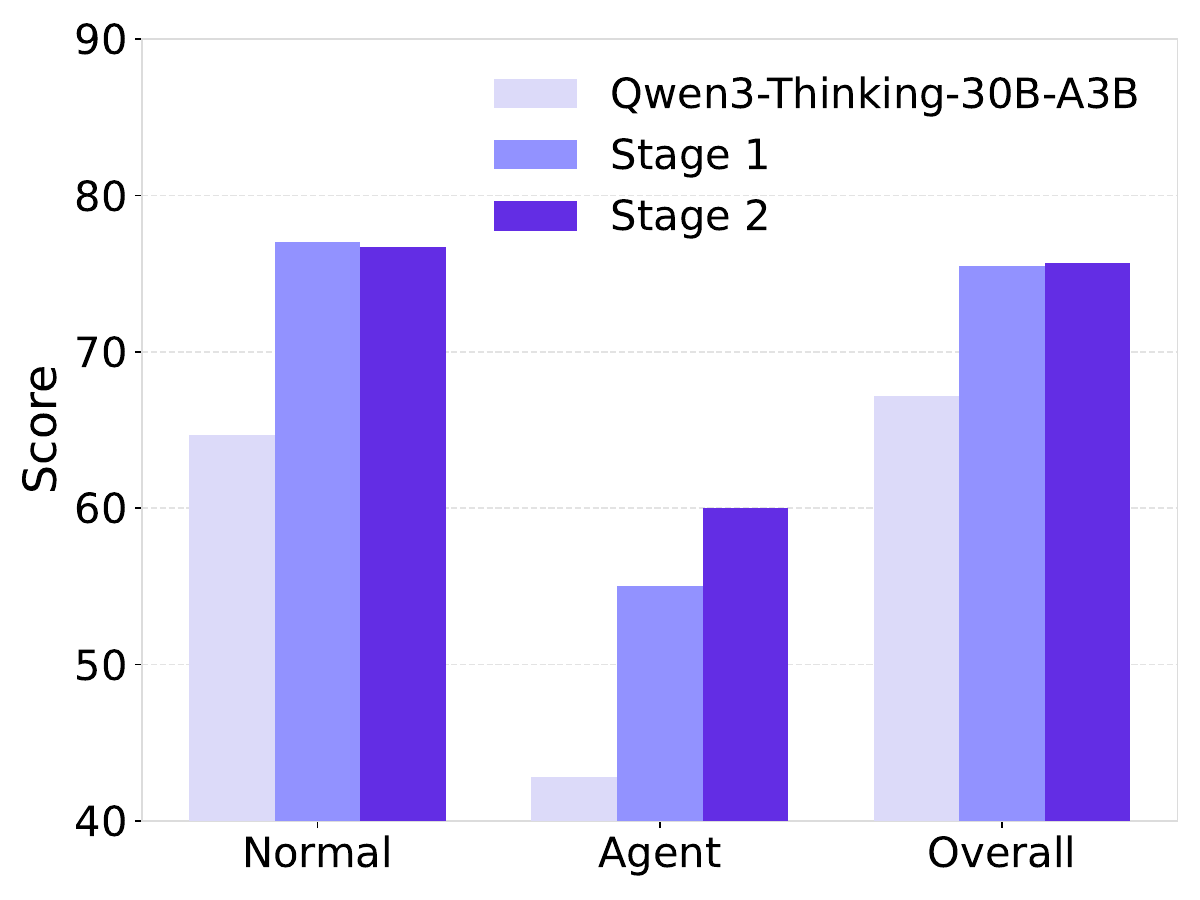}
\caption{Performance comparison on the \texttt{Normal}, \texttt{Agent}, and \texttt{Overall} subsets of ACEBench-en for two-stage training models.
}
\label{fig:stage_1_2_figure}
\vspace{-1mm}
\end{wrapfigure}
Notably, \textbf{AgentScaler‑4B} achieves performance on par with 30B‑parameter models despite using the fewest parameters, highlighting the agentic potential of compact LLMs.
Moreover, \textbf{AgentScaler-30B-A3B} delivers results that are comparable to trillion-parameter open-source models and, in several domains, approach those of closed-source counterparts.
These findings highlight the efficiency of our approach: agentic capabilities can be effectively learned and deployed even in relatively compact models, enabling competitive performance without relying on massive parameter counts. 
This advantage makes AgentScaler particularly well-suited for practical deployment in resource-constrained or latency-sensitive scenarios.

\paragraph{Ablation Study} We further conduct an ablation analysis to examine the effect of the proposed two-stage agent experience learning framework on ACEBench-en. As shown in Figure~\ref{fig:stage_1_2_figure}, both Stage~1 and Stage~2 training substantially improve performance over the base model (Qwen3-Thinking-30B-A3B) across all subsets. And through multi-steps agent training in Stage~2, the model's score on the agent set has further improved, and the overall score has also increased.
These results validate the design of the two-phase training pipeline: general foundation learning is critical for establishing tool-usage competence, and subsequent domain-specialization further consolidates and contextualizes these capabilities.

\section{Analysis}


\begin{wraptable}{r}{8cm}
\small
\centering
\caption{The \textbf{results} on ACEBench-zh.}
\begin{adjustbox}{width=1.0\linewidth}
\centering
\label{tab:acebench_zh}
\begin{tabular}{l|cccc}
\toprule
\multirow{2}{*}{\textbf{Model}}& \multicolumn{4}{c}{\textbf{ACEBench-zh}}     \\ 
& \texttt{Normal} & \texttt{Special} & \texttt{Agent} & \texttt{Overall} \\
\toprule
Qwen3-Thinking-4B & 34.7&85.3&6.7&43.9\\
AgentScaler-4B  & 70.8{\tiny\textcolor{red}{+36.1}} & 70.0{\tiny\textcolor{blue}{-15.3}}& 38.4{\tiny\textcolor{red}{+31.7}}	&65.6{\tiny\textcolor{red}{+21.7}} \\
\arrayrulecolor{black!20}\midrule
Qwen3-8B & 80.3& 72.7& 35.0& 71.3\\
AgentScaler-8B  & 75.2{\tiny\textcolor{blue}{-5.1}}& 79.3{\tiny\textcolor{red}{+6.6}}& 58.4{\tiny\textcolor{red}{+23.4}}	& 73.7{\tiny\textcolor{red}{+2.4}}\\
\arrayrulecolor{black!20}\midrule
Qwen3-Thinking-30B-A3B & 73.4& 86.7& 55.8& 74.2\\
AgentScaler-30B-A3B  & 85.3{\tiny\textcolor{red}{+11.9}}& 83.3{\tiny\textcolor{blue}{-3.4}}& 64.1{\tiny\textcolor{red}{+8.3}}& 81.5{\tiny\textcolor{red}{+7.3}}\\
\arrayrulecolor{black}\bottomrule
\end{tabular}
\end{adjustbox}

\end{wraptable}

\paragraph{Our synthetic data approach enables efficient knowledge transfer and strong robustness and generalization.}
We further evaluate our models on ACEBench-zh, which represents an out-of-distribution (OOD) scenario relative to the training setup. 
As shown in Table~\ref{tab:acebench_zh}, the AgentScaler models consistently outperform their Qwen baselines across all scales in terms of overall score. 
In particular, AgentScaler-30B-A3B achieves the best overall score of 81.5, demonstrating strong improvements in both the Normal and Agent subsets, while maintaining competitive performance on the Special subset. Notably, the small Qwen3-4B model demonstrated a remarkable improvement in agentic capabilities after the two-stage training, with its score surging from 6.7 to 38.4 and substantial gain of 21.7 points in the overall score. This offers valuable insights into effectively training compact models for complex function calling tasks in real-world applications.
\begin{figure}[t]
    \centering
    \includegraphics[width=0.99\textwidth]{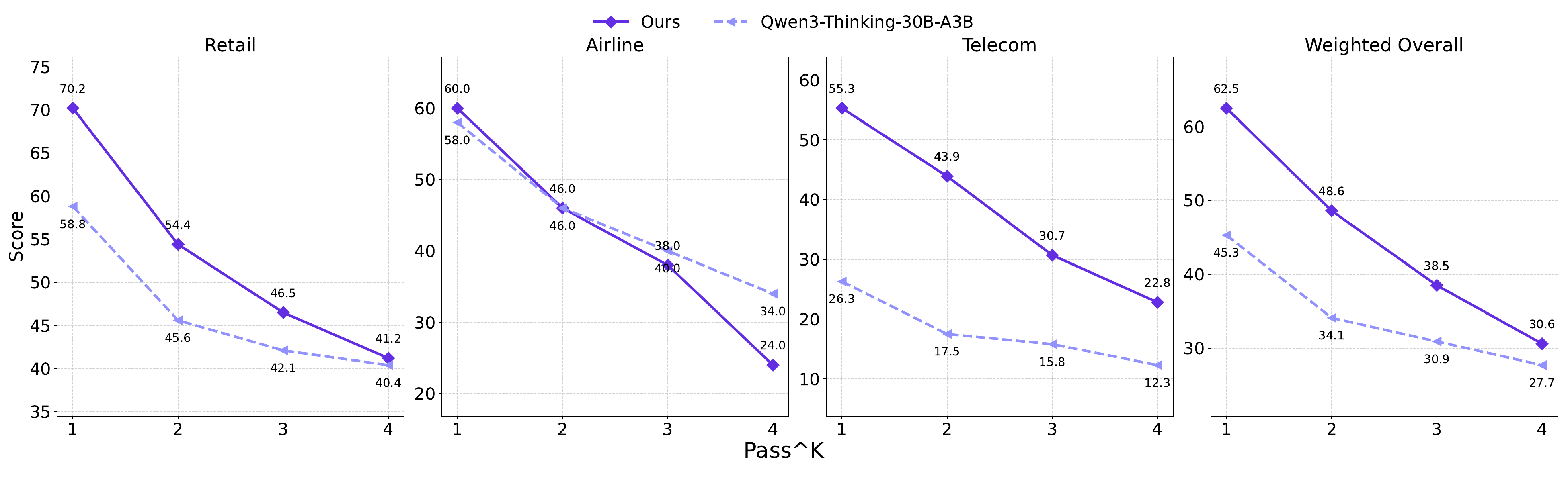}
    \caption{Pass$\textasciicircum$ k metric results across all domains in the $\tau^2$-Bench.}
    \label{fig:passk}
\end{figure}

\noindent \textbf{AgentScaler shows the strong consistency, stability.}
To assess the stability of AgentScaler, Figure~\ref{fig:passk} reports the pass$\textasciicircum$ k metric on the $\tau^2$-Bench, which denotes the accuracy achieved when the model correctly answers the same question in all k independent trials.
According to the experimental results, the weighted overall score of AgentScaler‑30B‑A3B consistently surpasses that of Qwen3‑Thinking‑30B‑A3B across all evaluated pass$\textasciicircum$ k settings, indicating a substantial performance advantage of our model over Qwen3‑Thinking‑30B‑A3B. 
Moreover, a clear downward trend in scores is observed as k increases, suggesting that the stability of existing LLMs remains a considerable challenge.


\begin{wrapfigure}{l}{11cm}
\vspace{-5mm}
\centering
\includegraphics[width=0.65\columnwidth]{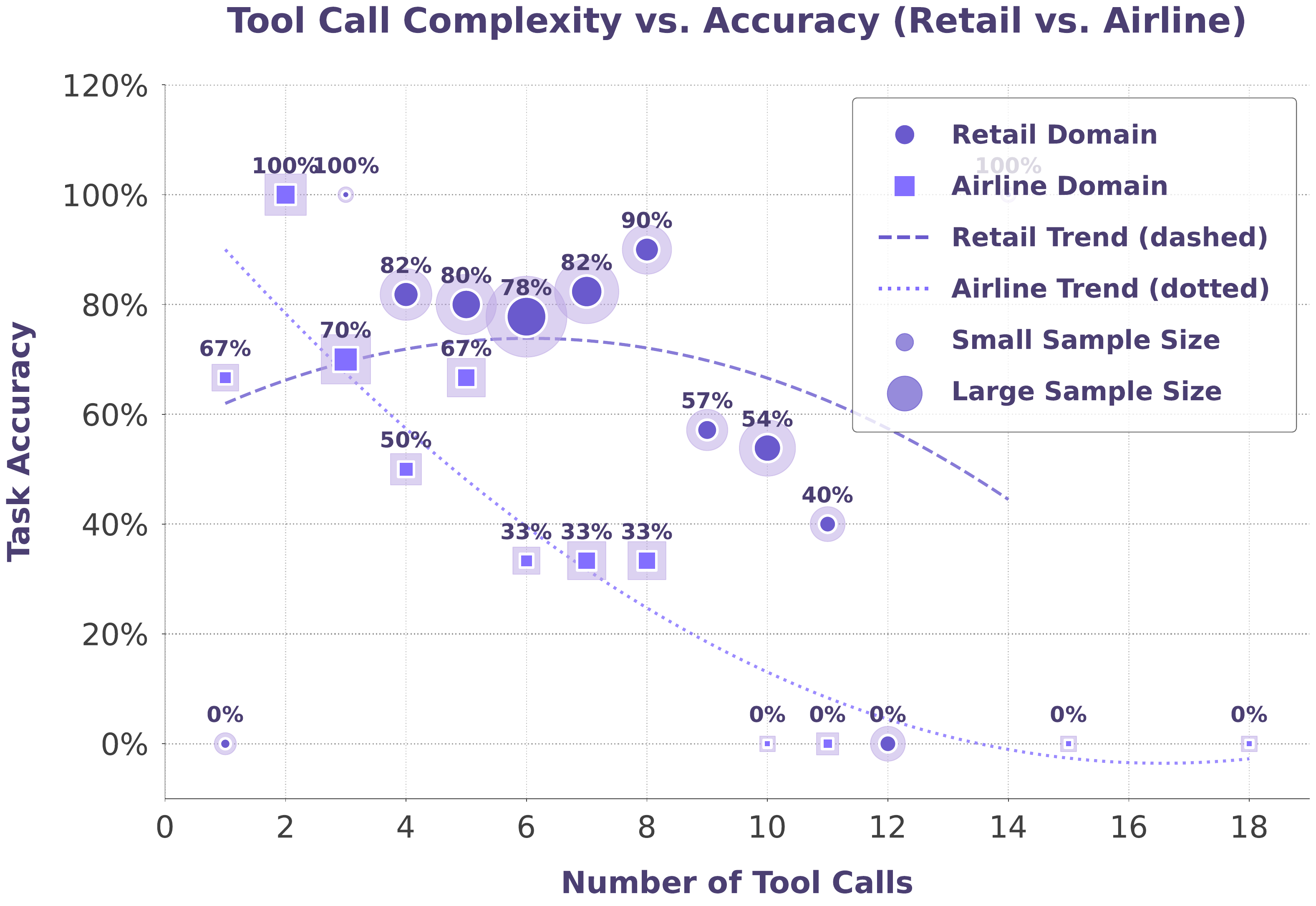}
\caption{Accuracy by tool call count on $\tau$-bench.}
\label{fig:domain_toolcall_acc}
\vspace{-5mm}
\end{wrapfigure}
\paragraph{Long-horizon tool calling remains a fundamental challenge for agentic models.} 
To further analyze the model’s long-horizon tool‑calling capability, we constructed a scatter plot on the $\tau$-bench dataset showing the relationship between the number of tool calls in each trajectory and the corresponding trajectory accuracy, with a dashed line indicating the trend.
As illustrated in Figure~\ref{fig:domain_toolcall_acc}, there exists a clear negative correlation between the number of tool calls and task accuracy. 
Our AgentScaler models exhibit this trend as well, underscoring that handling extended tool-use chains is still an open problem that we plan to address in future work.
\section{Related Work}
\subsection{Tool-Use Environments}
The construction of tool-use environments primarily involves three approaches: real-world environments, LLM--simulated environment, and Simulated Environments based on a state config. Using real-world environments~\citep{qin2023toolllm,song2023restgpt,mastouri2025making,wu2025webwalker} to invoke actual tools yields the most authentic feedback and enhances the model’s robustness in practical applications. However, this requires frequent calls to MCP services, resulting in high costs and significant time overhead. Moreover, maintaining a highly available and stable MCP service is often difficult, posing major challenges for agentic data generation and online RL training of models.Many works use LLM-generated responses to simulate environments as a source of tool responses~\citep{qin2024tool,lu2024toolsandbox,sun2025zerosearch}. By leveraging strong or fine-tuned LLMs, these approaches generate plausible responses given a tool call. However, such methods struggle with issues like hallucination and inconsistent response variability.
To address the limitations of the above two approaches, some recent work~\citep{ye2025feedback,taubench, tau2,prabhakar2025apigen,ye-etal-2025-toolhop} proposes building an offline tool execution environment for LLM training and evaluation. On one hand, offline environments avoid calling real tools, significantly reducing response generation cost and latency. On the other hand, mocked tool usage in such environments can still interact with real databases or state files through actual execution. However, these methods are more commonly applied in LLM evaluation rather than training, as constructing a reliable tool suite and a high-fidelity execution environment typically requires substantial manual effort. Furthermore, it is difficult to automatically validate the quality of such environments without human involvement, making scalability a significant challenge.
Our approach enables domain scalability through sampling from a toolgraph, and eliminates the need for human intervention via a rigorous, rule-based validation pipeline. This makes scalable construction of tool execution environments feasible.
\subsection{Tool Learning}
To enhance the agentic capabilities and tool-calling abilities of models, many works have attempted to improve tool utilization through various approaches. For instance, xLAMs~\citep{prabhakar2025apigen,zhang2024xlam} and ToolAce~\citep{liu2024toolace} leverage large-scale agentic data synthesis pipelines to generate high-quality training data and thereby boost model performance. DiaTool-DPO~\citep{jung2025diatool} employs DPO to enable models to learn from multi-turn positive and negative trajectories. Meanwhile, Tool-RL~\citep{qian2025toolrl}, Tool-N1~\citep{tooln1} utilize reinforcement learning (RL) algorithms to enhance both the tool-calling proficiency and generalization ability of models, further pushing the performance boundaries beyond supervised fine-tuning. Overall, whether relying on agentic data synthesis or online interaction with environments via RL training, a stable, reliable, and scalable execution environment is essential. For example, Kimi-K2~\citep{team2025kimi} uses a tool simulator during Agentic Data Synthesis to obtain observations for multi-turn trajectories. Our method not only leverages accurately simulated tool environments to collect trajectories but also introduces verifiable environmental state changes, making each simulation response more reliable. Furthermore, we propose a state change based environment validation strategy, enabling a robust filtering mechanism for large-scale agentic data synthesis.
\section{Conclusion}
In this work, we presented a principled pipeline for advancing general agentic intelligence through systematic environment scaling and agent experience learning. By programmatically materializing tools as executable code and grounding them in database-structured environments, our approach enables large-scale construction of verifiable trajectories. Building on these environments, we introduced a two-stage agent experience learning framework that first equips agents with fundamental tool-usage capabilities and then specializes them for domain-specific contexts.
Extensive experiments on three representative benchmarks, $\tau$-bench, $\tau^2$-Bench, and ACEBench, demonstrate the effectiveness of our pipeline. Notably, our AgentScaler family achieves state-of-the-art performance among open-source models under 1T parameters, and in several cases reaches parity with much larger or closed-source counterparts.

Looking ahead, we believe our work highlights the importance of scalable environment construction and verifiable agentic experience for fostering robust and generalizable language agents. 
Future directions include integrating reinforcement learning on top of our fully simulated environments and extending our pipeline to broader modalities and real-world deployment scenarios.
\section*{Limitation}
Although our proposed framework has demonstrated promising results, several limitations remain, which point to ongoing efforts and potential directions for future work.
\paragraph{Lack of Reinforcement Learning} While our approach leverages two-stage supervised fine-tuning (SFT), \textbf{the simulated environment we construct provides stable and low-latency feedback, which is inherently well-suited for RL optimization}. 
In future work, we aim to integrate RL on top of our simulated environment to further improve the agentic behavior of the model.
\paragraph{Model Scale} Another limitation of our current work lies in model scale. 
Our method has so far only been validated on a 30B-scale architecture, without extension to larger models exceeding 200B or even trillion-parameter scales.
While prior work~\citep{belcak2025small} emphasizes that ``small language models are the future of agentic AI,'' we share the view that training agentic capabilities in relatively smaller models is particularly meaningful.
Such models are easier to deploy on edge devices, enable broader applicability across diverse scenarios, and offer faster response times.
\clearpage

\clearpage
\bibliography{biblio}
\bibliographystyle{colm2024_conference}

\clearpage
\appendix

\end{document}